\documentclass[letterpaper, 10 pt, conference]{ieeeconf}
\IEEEoverridecommandlockouts                              
\usepackage{graphicx} 
\usepackage{amsmath} 
\usepackage{amssymb}  

\providecommand*{\ordinarycolon}{:}
\providecommand*{\vcentcolon}{\mathrel{\mathop\ordinarycolon}}
\providecommand*{\coloneqq}{\vcentcolon\mathrel{\mkern-1.2mu}=}
\providecommand*{\R}{\mathbb R}
\providecommand*{\skewprod}{\null\mathord\times}
\usepackage{microtype}

\usepackage{color}
\usepackage{cancel}
\usepackage[english]{babel}

\title{\LARGE \bf
Lightweight Model Predictive Control for\\
Spacecraft Rendezvous Attitude Synchronization
}

\author{Peter Stadler\textsuperscript{1}, Alexander Meinert\textsuperscript{1}, Niklas Baldauf\textsuperscript{1} and Alen Turnwald\textsuperscript{2}
\thanks{
This work was supported by the Bavarian Ministry of Economic Affairs,
Regional Development and Energy (grant no. MRF-2307-0009)}
\thanks{\textsuperscript{1}Peter Stadler, Alexander Meinert and Niklas Baldauf are with the Space Applications Group, e:fs TechHub GmbH, 85080 Gaimersheim, Germany {\tt\{Peter.Stadler, Alexander.Meinert, Niklas.Baldauf\}@\allowbreak{}efs-techhub.com}}
\thanks{\textsuperscript{2}Alen Turnwald is with the Faculty of Electrical Engineering and Information Technology, Ingolstadt University of Applied Sciences, Germany {\tt Alen.Turnwald@thi.de}}
}

\begin{document}

\maketitle
\thispagestyle{empty}
\pagestyle{empty}

\begin{abstract}
This work introduces two lightweight model predictive control~(MPC) approaches for attitude tracking with reaction wheels during spacecraft rendezvous synchronization.
Both approaches are based on a novel attitude deviation formulation, which enables the use of inherently linear constraints on angular velocity.
We develop a single-loop and a dual-loop MPC; the latter embeds a stabilizing feedback controller within the inner loop, yielding a linear time-invariant system.
Both controllers are implemented with CasADi---including automatic code generation---evaluated across various solvers, and validated within the Basilisk astrodynamics simulation framework.
The experimental results demonstrate improved tracking accuracy alongside reductions in computational effort and memory consumption.
Finally, embedded delivery to an ARM Cortex-M7---representative of commercial off-the-shelf devices used in New Space platforms---confirms the real-time feasibility of these approaches and highlights their suitability for onboard attitude control in resource-constrained spacecraft rendezvous missions.
\end{abstract}

\section{\textsc{Introduction}}

Attitude pointing and tracking are standard spacecraft operations.
A challenging example arises during the rendezvous of two satellites, where attitude is controlled using reaction wheels.
We focus on the synchronization of rotational motion as described in~\cite{Colmenarejo2018}.
During this phase, the servicer aligns its angular velocity to the target’s momentum vector and tracks its attitude.
The stabilization phase would involve a similar task: detumbling amounts to attitude pointing.
The difficulty in this phase arises from the uncertain inertia of the coupled system, for which~\cite{Lee2011} proposes a Lyapunov controller.

There are various ways to represent attitude.
Rotation matrices describe the attitude continuously and uniquely, and we use them for certain purposes as they yield compact formulas.
Unit quaternions are a common representation used for attitude tracking as in~\cite{VanDyke}.
However, neither is a minimal representation of the inherently $3$-dimensional attitude.

Projecting the quaternions down to generalized Rodrigues parameters, as in~\cite{Schaub1996}, introduces a singularity.
Instead, we lift the rotation matrices to rotation vectors in $\R^3$.
Rotation vectors correspond to elements of the Lie algebra $\mathfrak{so}(3)$ as used by~\cite{Teng2022} for developing an iterative linear-quadratic regulator (iLQR).

One disadvantage of rotation vectors is that they cannot represent orientations both uniquely and continuously.
This is not a drawback for attitude control, since no globally asymptotically stable continuous attitude controller exists~\cite{Kalabic2017}.

Enforcing safety and hardware constraints, such as angular velocity bounds and reaction wheel torque limitations, is paramount for attitude control.
Non-linear model predictive control~(MPC) considers these constraints explicitly like in~\cite{Kalabic2017} but demands high computational resources.
To address limited onboard resources, we leverage linear MPC, for which a variety of quadratic-program solvers are available.

For attitude tracking, the system can be linearized around the reference trajectory using the tracking error.
We introduce a novel multiplicative attitude-error definition obtained by reversing the conventional order of the rotation matrices.
This yields inherently linear angular-velocity constraints.

Based on this formulation, we introduce two linear MPCs for spacecraft attitude synchronization:
a conventional single-loop design, and a dual-loop design that uses a stabilizing feedback controller for the inner control loop.
The latter augments the dual-loop MPC in~\cite{Turnwald2023}, where the attitude error is the difference of unit quaternions.

\begin{figure}[t]\centering
\vskip+1.4ex
\!\includegraphics[width=\dimexpr\linewidth-0.2em]{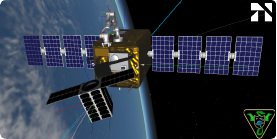}
\vskip-3ex
\end{figure}

High-fidelity simulations in Basilisk~\cite{Basilisk} validate improved attitude tracking performance of the proposed controllers compared to~\cite{Turnwald2023}.
We inspect different solvers for the quadratic programs given by the linear MPCs using CasADi~\cite{Andersson2018} and deliver the MPCs to an embedded commercial off-the-shelf processor.
The dual-loop MPC outperforms the other two linear MPCs in terms of runtime and memory usage.

\section{\textsc{Preliminaries}}

\subsection{Attitude Kinematics}

The attitude of a spacecraft is given by its rotation with respect to an inertial frame.
It can be represented by the direction cosine matrix $R\in\R^{3\times3}$.
The angular velocity of the spacecraft with respect to the inertial frame expressed in the spacecraft body frame is denoted as $\omega\in\R^3$.

The derivative of the rotation matrix obeys the kinematic equation:
\begin{equation}
\label{eq:kinematic}
\dot R = R\,\omega\skewprod
\end{equation}
Here, the operator $\skewprod$ maps a vector to its skew-symmetric matrix:
\begin{equation*}\textstyle
\left(\begin{smallmatrix}
x\\y\\z
\end{smallmatrix}\right)\mkern-5mu\skewprod
\coloneqq
\left(\begin{smallmatrix}
\hphantom{-}0 & \hphantom{}-z & \hphantom{-}y \\
\hphantom{-}z & \hphantom{-}0 & \hphantom{}-x \\
\hphantom{}-y & \hphantom{-}x & \hphantom{-}0
\end{smallmatrix}\right)
\end{equation*}
It yields the cross-product by multiplying with another vector.

\subsection{Rotation Vector}

For optimization purposes as in MPC, it is desirable to represent the attitude in a $3$-dimensional vector space: solvers can use it without additional constraints or projections.
We lift the direction cosine matrices to rotation vectors, which have this property.
A rotation vector $\phi\in\R^3$ is co-directional with the axis and its length $\lVert\phi\rVert_2$ equals the angle of the rotation.
The matrix exponential gives $\exp(\phi\skewprod) = R$.

The theoretical superstructure is that the rotation matrices form the Lie group $\mathrm{SO} (3)$.
They are constrained to have the determinant $\det(R) = 1$.
The corresponding Lie algebra $\mathfrak{so}(3)$ forms a vector space (essentially the rotation vectors); this is the key benefit.
Other advantages of using the Lie algebra for control are elaborated in \cite{Teng2022}.

The advantages of the rotation matrices are that composing rotations corresponds to multiplying matrices, and that the kinematics~(\ref{eq:kinematic}) are easier to handle. The kinematics in terms of the rotation vector result in the non-linear Bortz equation~\cite{Bortz}:
\begin{align}
\label{eq:bortz}
\dot\phi = \omega + \phi\times\left(\frac12\,\omega + \frac1{\lVert\phi\rVert_2^2}\left(1 - \frac{\lVert\phi\rVert_2\,\sin\lVert\phi\rVert_2}{2 - 2\,\cos\lVert\phi\rVert_2}\right)\phi\times\omega\right)
\end{align}

\subsection{Attitude Dynamics}

The angular velocity $\omega$ obeys the dynamic equation:
\begin{align}
\label{eq:dynamic}
J\,\dot\omega = \tau - \omega\times L
\end{align}
Here, $\tau$ is the applied torque, $J$ is the inertia tensor of the spacecraft, and $L$ is its angular momentum.

For rigid bodies, the angular momentum would be $J\,\omega$; however, the spacecraft considered here has moving parts due to the use of reaction wheels.
Actuating reaction wheels or other momentum-exchange devices keeps the total angular momentum $L$ constant (law of conservation), and the angular acceleration $\dot\omega$ depends linearly on the torque $\tau$ and the current angular velocity $\omega$.
This is valid only for momentum-exchange devices, not for actuators that generate an external torque, such as attitude thrusters and magnetorquers.

\section{\textsc{Methods}}

For attitude tracking, we focus on controlling the spacecraft to follow a desired trajectory given in a reference frame by:
\begin{itemize}
\item rotation matrix $R_d$,
\item angular velocity $\omega_d$ and
\item angular acceleration $\dot\omega_d$
\end{itemize}
These reference functions depend on time and fulfill the kinematic equation $\dot R_d = R_d\,\omega_d\skewprod$, analogous to~(\ref{eq:kinematic}).

To constrain the components of the exerted torque $\tau$ and of the spacecraft’s angular velocity $\omega$, we assume some realistic lower bounds $\tau_-$ and $\omega_-$ as well as upper bounds $\tau_+$ and~$\omega_+$.

\subsection{Proposed Error System}\label{sec:error_system}

We define the attitude deviation in reverse order:
\begin{align}
\nonumber
\Delta R &\coloneqq R\,R_d^T
\intertext{and the deviation of the angular velocity as:}
\label{eq:delta_omega}
\Delta\omega &\coloneqq R_d\,(\omega - \omega_d)
\end{align}

It is common to define the attitude deviation as $R_d^T\,R$ or the same quantity in another representation, e.g., as rotation vector~\cite{Teng2022}.
The reversed deviation has advantages for MPC as it yields linear angular-velocity constraints~(\ref{eq:single_loop_omega}) and~(\ref{eq:dual_loop_omega}).

Although the error in reverse order lacks the usual interpretation as a coordinate transformation, the product is mathematically valid and equals the identity $I_3$ if and only if $R_d=R$.
We rely purely on the mathematical properties of this reversed matrix product.
The deviations fulfill an equation similar to the kinematics in~(\ref{eq:kinematic}):
\begin{align*}
\Delta \dot{R} = R\,(\omega\skewprod - \omega_d\skewprod)\,R_d^T = \Delta R\,\Delta\omega\skewprod
\end{align*}
This follows from the identity
$\left(R_d^T\,\Delta\omega\right)\skewprod = R_d^T\,\Delta\omega\skewprod\,R_d$.

A rotation vector error $\Delta\phi$ corresponding to $\Delta R$ obeys the respective Bortz equation~(\ref{eq:bortz}).
The vector is not uniquely determined because of its $2\,\pi$-periodicity.
We could restrict its length to $\pi$ using the main branch of the matrix logarithm of $\Delta R$.
However, since we recompute $\Delta\phi$ at each control step, its sign could flip at $\pi$.
Instead, we relax this restriction to make the calculation continuous for small steps.

\subsection{Single-Loop Linear MPC}\label{sec:single_loop_mpc}

\begin{figure}[thbp]\centering
\includegraphics[width=\dimexpr\linewidth-0em]{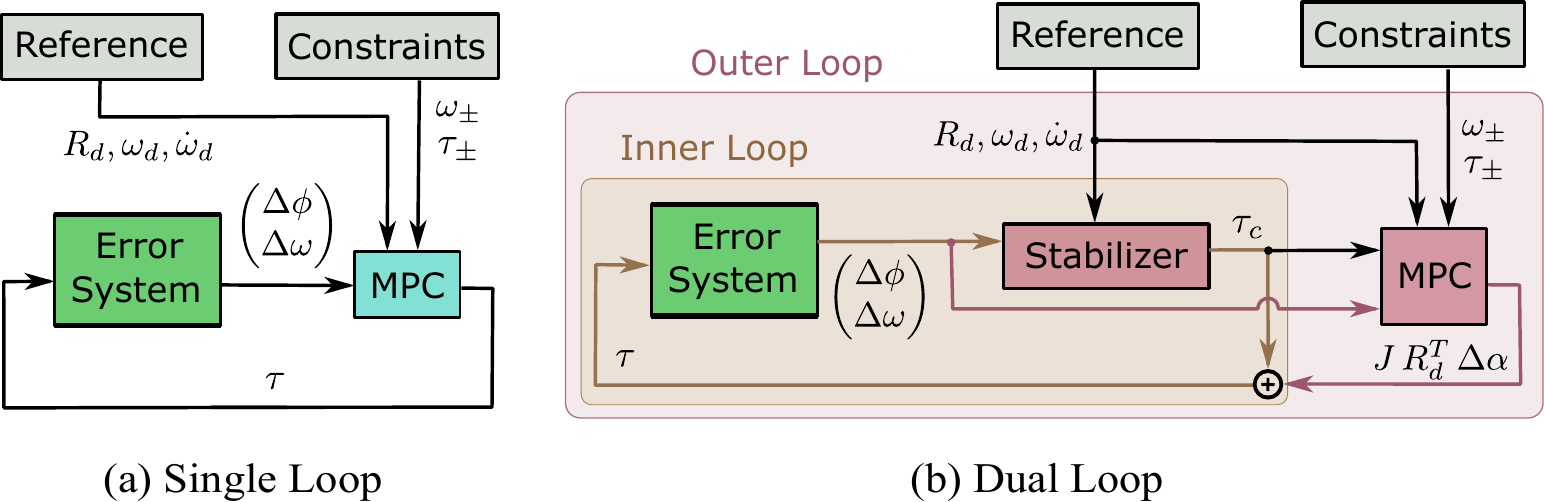}\vskip-1ex
\caption{Layouts of the Proposed Linear MPCs for Attitude Tracking
\\
Both designs use the same reference trajectory, the same constraints, and the proposed error system based on the attitude deviation in reverse order.
\\
(a)\hskip.5em The single-loop MPC~(\ref{eq:single_loop_mpc}) directly uses the error system linearized at zero:
The model for the predictions is a discretization of the affine but parameter-varying~(LPV) error system~(\ref{eq:linear_error_system}).
\\
(b)\hskip.5em The dual-loop MPC~(\ref{eq:dual_loop_mpc}) linearizes and discretizes an already stable inner loop for the MPC in the outer loop.
The inner system is the closed loop of the error system with the stable feedback controller~(\ref{eq:stable_control_torque}), which also eliminates the parameter variance. The linearization of the inner loop results in the stable and time-invariant~(LTI) system~(\ref{eq:closed_loop_system}).
}\footnotesize
\label{fig:layouts}
\end{figure}

Linearizing the Bortz equation of the error system (at 0) and differentiating the angular velocity error $\Delta\omega$ leads to the parameter-varying~(LPV) error system:
\begin{subequations}
\label{eq:linear_error_system}
\begin{align}
\Delta\dot{\phi} &\approx \Delta\omega
\\
\Delta\dot{\omega} &= R_d\,(\omega_d\times\omega + \dot\omega - \dot\omega_d)
\end{align}
\end{subequations}
The latter is affine in $\tau$ and $\Delta\omega$, since $\dot\omega$ depends linearly on $\tau$ and $\omega$ according to~(\ref{eq:dynamic}), and $\omega$ is an affine transform of $\Delta\omega$ because of~(\ref{eq:delta_omega}).
The linearized error system is used as model for the single-loop MPC (Figure \ref{fig:layouts}a).

The MPC has a finite horizon of $N$ constant increments $\Delta t$, a state cost matrix $Q_\textrm{single-loop}$ and an input cost matrix $R_\textrm{single-loop}$ (both  symmetric and positive semi-definite). The terminal cost matrix $P_\textrm{single-loop}$ and positively invariant polytope $X_\textrm{single-loop}$ are calculated by solving the discrete algebraic Riccati equation.
For time-dependent quantities, the index $k$ indicates its value at time $k\, \Delta t$ in the future; $x_0$ estimates the current state, and the other $x_k$ denote the predicted states:
\begin{align}
\label{eq:state_vector}
x_k&\coloneqq
\begin{pmatrix}
\Delta\phi_k
\\
\Delta\omega_k
\end{pmatrix}
=
\begin{pmatrix}
\Delta\phi(k\,\Delta t)
\\
\Delta\omega(k\,\Delta t)
\end{pmatrix}
\end{align}

The single-loop linear MPC is formulated as the optimal control problem by the following quadratic program (QP):
\begin{subequations}
\label{eq:single_loop_mpc}
{
\setlength\belowdisplayskip{0pt plus 0pt minus 0pt}
\setlength\belowdisplayshortskip{0pt plus 0pt minus 0pt}
\begin{equation}
\min_{\tau_k}
\left\lVert x_N \right\rVert_{P_\textrm{single-loop}}^2 +
\sum_{k<N}
\left\lVert x_k \right\rVert_{Q_\textrm{single-loop}}^2 + \left\lVert\tau_k\right\rVert_{R_\textrm{single-loop}}^2
\end{equation}
}
{
\setlength\abovedisplayskip{\jot}
\setlength\abovedisplayshortskip{\jot}
\begin{align}
\text{subject to }\quad
x_{k+1} &=  A_k\,x_k + B_k\,\tau_k + c_k
\\
x_N &\in X_\textrm{single-loop}
\\
\omega_k
&\in [\omega_-,\omega_+]^3
\\
\tau_k &\in [\tau_-, \tau_+]^3
\intertext{The system matrices $A_k$, $B_k$, and the offset vector $c_k$ discretize the affine but parameter-varying~(LPV) error system~(\ref{eq:linear_error_system}) by zero order hold.
The predicted angular velocity $\omega_k$ is affine in the optimization variable $\Delta\omega_k$:}
\omega_k &\coloneqq \omega_{d,k} + R^T_{d,k}\,\Delta\omega_k \label{eq:single_loop_omega}
\end{align}
}
\end{subequations}

The angular velocity $\omega_k$ is affine since we defined the attitude error $\Delta R$ in reverse order in~(\ref{eq:delta_omega}).
Using the conventional attitude-error definition, the angular-velocity deviation would be $\omega - R^TR_d\,\omega_d$, where the rotation matrix depends nonlinearly on the rotation vector.
If this conventional angular-velocity error were linearized to preserve linear constraints on angular velocity, prediction accuracy would deteriorate.

\subsection{Stable Feedback Controller}\label{sec:stable_controller}

We define a stabilizing control torque for attitude tracking similar to~\cite{VanDyke}, but use rotation vectors instead of quaternions:
\begin{align}
\label{eq:stable_control_torque}
\tau_c \coloneqq \underbrace{\omega\times L}_\text{gyroscopic} + \underbrace{J\,\left(\dot\omega_d - \omega_d\times\omega\right)}_\text{tracking feedforward} - \underbrace{J\,R_d^T\,\left(k_c\,\Delta\phi + k_\omega\,\Delta\omega\right)}_\text{linear feedback}
\end{align}
The positive gains $k_c$ and $k_\omega$ can be chosen arbitrarily.

The control law $\tau_c$ serves as the stabilizer of the inner loop in Figure~\ref{fig:layouts}b. Together with the error system~(\ref{eq:linear_error_system}), it forms the inner loop of the dual-loop MPC defined below.
It is designed to simplify the derivative of the angular velocity error:
\begin{align}
\label{eq:closed_domega}
\Delta\dot{\omega} = -k_c\, \Delta\phi - k_\omega\,\Delta\omega
\end{align}

Stability of the controller can be shown with the function:
\begin{align*}
V\coloneqq \frac{k_c}2\,\lVert\Delta\phi\rVert_2^2 + \frac12\,\lVert\Delta\omega\rVert_2^2
\end{align*}

$V$ is a Lyapunov function: it is clearly positive definite and radially unbounded; its derivative is non-positive by~(\ref{eq:bortz}),~(\ref{eq:closed_domega}), and the fact that the product $\Delta\phi\times ...$ is orthogonal to $\Delta\phi$:
\begin{align*}
\dot V
 & = k_c\, \langle\Delta\phi\mid\Delta\dot{\phi}\rangle +\langle\Delta\omega\mid\Delta\dot{\omega}\rangle \\
 &= k_c\, \langle\Delta\phi\mid\cancel{\Delta\omega}
 + \Delta\phi\times ...\rangle
 + \langle\Delta\omega\mid\cancel{-k_c\, \Delta\phi} - k_\omega\,\Delta\omega\rangle \\
&= -k_\omega\,\lVert\Delta\omega\rVert_2^2
\end{align*}

By~(\ref{eq:closed_domega}), the equilibrium point at the origin $0$ is the only trajectory with constant $\Delta\omega\equiv0$.
LaSalle's invariance principle~\cite{LaSalle} implies global asymptotic stability.
Similar to~\cite{Lee2011}, one could prove even exponential stability by adding $c\,\langle\Delta\phi\mid\Delta\omega\rangle$ to $V$ and determining $c$ for $\lVert\Delta\phi\rVert \le M <2\,\pi$.

\subsection{Inner Loop System}\label{sec:closed_loop_system}

Closing the loop around the error system~(\ref{eq:linear_error_system}) yields a parameter- and time-invariant~(LTI) system:
\begin{align}
\label{eq:closed_loop_system}
\frac{\mathrm d}{\mathrm dt}
\begin{pmatrix}
\Delta\phi \\ \Delta\omega
\end{pmatrix}
\approx
\begin{pmatrix}
0 & I_3 \\
-k_c\,I_3 & -k_\omega\,I_3
\end{pmatrix}
\begin{pmatrix}
\Delta\phi \\ \Delta\omega
\end{pmatrix}
+
\begin{pmatrix}
0 \\ I_3
\end{pmatrix}
\Delta\alpha
\end{align}
The system remains parameter-invariant if the input acceleration $\Delta\alpha$ is added linearly, via $J\,R_d^T\,\Delta\alpha$, to the stabilizer~(\ref{eq:stable_control_torque}).
It models the inner loop of the dual-loop MPC (Figure~\ref{fig:layouts}b).

The system has the poles:
\begin{equation*}
 - \frac{k_c + k_\omega}2 \pm \sqrt{\frac{k_\omega^2}4 - k_c}
\end{equation*}
It is asymptotically stable for all positive gains. We set the discriminant to zero by choosing
$4\,k_c = k_\omega^2$.
Our experiments suggest satisfactory performance for attitude tracking, at least for $0.5<k_\omega<2$.

\begin{figure*}[thbp]\centering
\includegraphics[width=\dimexpr\linewidth]{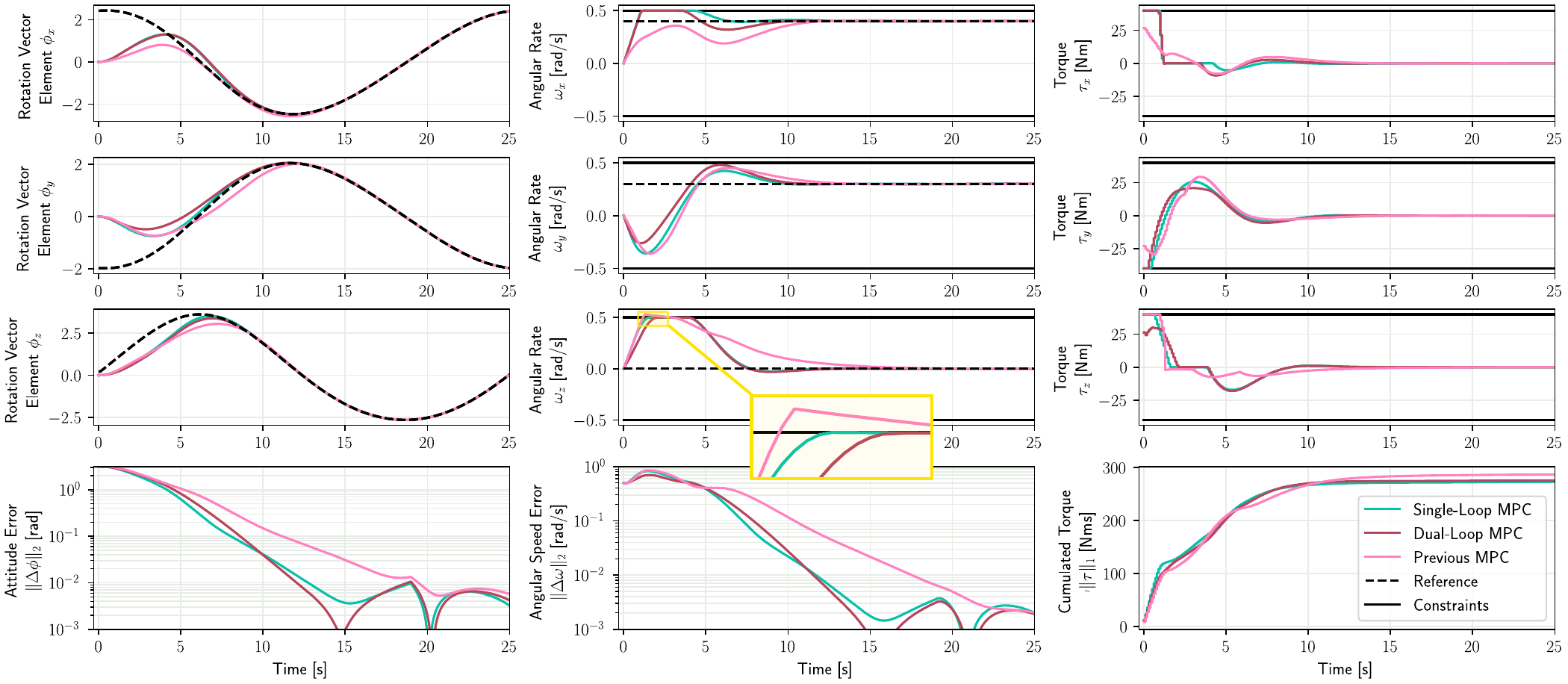}\vskip-.5ex
\caption{Attitude Tracking Quality of Different MPCs in Simulation with Basilisk
\\
The proposed dual-loop MPC and the single-loop MPC are quite similar and better than the previous MPC from~\cite{Turnwald2023}.
For the former two MPCs, the deviations of both attitude and angular speed fall faster with time and the effort measured as accumulated torque is slightly lower.
The angular velocity error for the previous MPC must be linearized.
This makes it difficult to constrain the angular velocity exactly; here, $\omega_z$ exceeds the upper bound (highlighted).
}
\label{fig:quality}
\end{figure*}

\subsection{Dual-Loop Linear MPC}

This controller uses the already stable inner system~(\ref{eq:closed_loop_system}) as model (Figure~\ref{fig:layouts}b). It connects the stabilizer~(\ref{eq:stable_control_torque}) with the error system~(\ref{eq:linear_error_system}); the single-loop MPC uses the latter solely.

Again, the MPC has a horizon of $N$ steps of length $\Delta t$, a state cost matrix $Q_\textrm{dual-loop}$ and an input cost matrix $R_\textrm{dual-loop}$  (both symmetric and positive semi-definite).
The terminal cost matrix $P_\textrm{dual-loop}$ and positively invariant polytope $X_\textrm{dual-loop}$ are calculated by solving the discrete algebraic Riccati equation.
The index $k$ denotes the discretization at $k\, \Delta t$ in the future, $x_0$ the current estimation and $x_k$ the predicted states (\ref{eq:state_vector}).

The dual-loop linear MPC is formulated as the following quadratic program~(QP). The terms shared with~(\ref{eq:single_loop_mpc}) are {\color[gray]{0.4}gray}:~
\begin{subequations}
\label{eq:dual_loop_mpc}
{
\setlength\belowdisplayskip{0pt plus 0pt minus 0pt}
\setlength\belowdisplayshortskip{0pt plus 0pt minus 0pt}
\begin{equation}
\min_{\Delta\alpha_k\vphantom{A^A}}
\left\lVert
x_N
\right\rVert^2_{P_\textrm{dual-loop}}
+
\sum_{k<N}
\left\lVert
x_k
\right\rVert^2_{Q_\textrm{dual-loop}}
+ \left\lVert\Delta\alpha_k\right\rVert^2_{R_\textrm{dual-loop}}
\end{equation}
}
{
\setlength\abovedisplayskip{\jot}
\setlength\abovedisplayshortskip{\jot}
\begin{align}
\text{subject to \quad}
x_{k+1}
& = A\,x_k + B\,\Delta\alpha_k
\\
x_N &\in X_\textrm{dual-loop}
\\
\color[gray]{0.4} \omega_k
&\color[gray]{0.4} \in [\omega_-,\omega_+]^3
\\
\color[gray]{0.4} \tau_k
&\color[gray]{0.4} \in [\tau_-, \tau_+]^3
\intertext{The system matrices $A$ and $B$ are constant as they discretize the parameter- and time-invariant~(LTI) system~(\ref{eq:closed_loop_system}) by zero order hold.
The predicted angular velocity $\omega_k$ and torque $\tau_k$ are affine relative to the optimization variables $x_k$ and $\Delta\alpha_k$:}
\color[gray]{0.4} \omega_k
&\color[gray]{0.4} \coloneqq \omega_{d,k} + R^T_{d,k}\,\Delta\omega_k\label{eq:dual_loop_omega}
\\
\tau_k
& \coloneqq \tau_{c,k} + J\,R^T_{d,k}\,\Delta\alpha_k
\end{align}
}
\end{subequations}

The constraints are linear: the torque $\tau_{c,k}$ of the stable controller~(\ref{eq:stable_control_torque}) is an affine function of $\Delta\phi_k$, $\Delta\omega_k$ and $\omega_k$; the angular velocity $\omega_k$ is affine, identical to~(\ref{eq:single_loop_omega}).

\section{\textsc{Results}}

We compare the two proposed MPCs to the dual-loop MPC from~\cite{Turnwald2023} in a simulated environment and analyze their runtime for increasing horizons.
The runtimes are visualized for each horizon with a violin plot on a logarithmic scale as they cover a wide range. Their geometric means are connected and the maximum values are emphasized.

\subsection{Comparison in Simulation}\label{sec:comparison}

We tuned the parameters to match the accuracy of the previous MPC given in~\cite{Turnwald2023}.
For the single-loop MPC~(\ref{eq:single_loop_mpc}) the selected cost matrices are:
\begin{align*}
Q_\textrm{single-loop}
=
\begin{pmatrix}
2\,I_3 & 0
\\
0 & 3\,I_3
\end{pmatrix}
\text{ and }
R_\textrm{single-loop}
=
5\,R_d\,J^{-1}
\end{align*}
For the dual-loop MPC~(\ref{eq:dual_loop_mpc}), the costs and  gain are set to:
\begin{align*}
Q_\textrm{dual-loop}
=
\begin{pmatrix}
10\,I_3 & 0
\\
0 & I_3
\end{pmatrix},
R_\textrm{dual-loop}
=
100\,I_3
\text{ and }k_\omega = 1.2
\end{align*}
This determines the other gain ($k_c=0.36$) as well as the terminal cost matrices and maximal positively invariant sets.

We set up the astrodynamics simulation for a spacecraft with reaction wheels in Basilisk~\cite{Basilisk}.
Every MPC was run with the same constraints, the same initial values, and the same fixed prediction time step. We used the following parameters:
\begin{itemize}
\item inertia tensor $J = \mathrm{diag}(85, 94, 92)\,\mathrm{kg \hskip.05em m^2}$,
\item reference angular velocity $\omega_d \equiv \bigl(0.4, 0.3, 0\bigr)^T\,\mathrm{rad}/\mathrm{s}$,
\item angular velocity constraints $\omega_\pm = \pm 0.5\,\mathrm{rad}/\mathrm{s}$,
\item torque constraints $\tau_\pm = \pm 40\,\mathrm{N\hskip.05em m}$, and
\item time increment $\Delta t = 0.1\,\mathrm{s}$.
\end{itemize}

To compare the quality of attitude tracking, the three MPCs run with a receding horizon of $N=11$ steps for $25$ seconds:
The simulation calls a controller with the current state, uses the first calculated torque $\tau_0$ as input for $\Delta t = 0.1\,\mathrm{s}$, and reinvokes the controller with the new state, closing the loop.
Figure~\ref{fig:quality} shows the advantages of the proposed two MPCs over the previous MPC:
inherently linear constraints, faster error decay and a slightly lower total torque requirement.

\begin{figure}[thbp]\centering
\includegraphics[width=\dimexpr\linewidth]{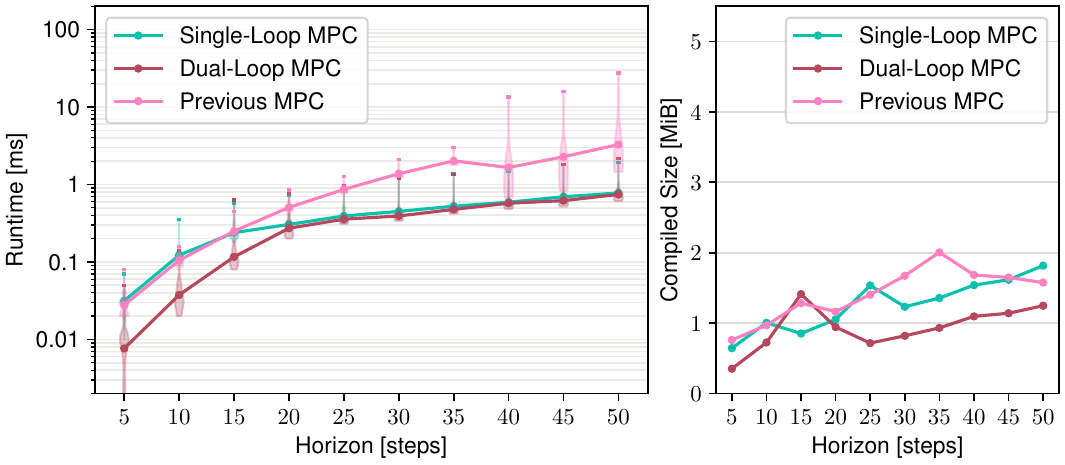}\vskip-1ex
\caption{Workload of Different MPCs in C++ on Intel i7-11850H
\\
The proposed dual-loop MPC is faster than the previous MPC.
For horizons less than 20 steps, the single-loop MPC is as slow as the previous MPC.
For larger horizons, the runtimes of the single-loop MPC and the proposed dual-loop MPC are similarly low, while the compiled size of the dual-loop MPC is smaller.
\\
For each algorithm and horizon, we choose the solver and sparsity pattern yielding the best runtime. The solver switches at 20 steps for the proposed dual-loop MPC and at 40 steps for previous MPC yield visible changes in runtimes distribution and compiled size.}
\label{fig:algorithms}
\end{figure}

For workload comparison of the three MPCs, we replay (in open loop) a run with a large horizon (high quality) for each MPC using horizons with different numbers of steps $N$. Figure~\ref{fig:algorithms} shows the advantages of the dual-loop MPC over the other two and that the single-loop MPC scales better than the previous MPC.
As it shows the best performance, we present the details for the dual-loop MPC in the following.

\subsection{Evaluation of Solvers}\label{sec:solvers}

\begin{figure}[b]\centering
\!\includegraphics[width=\dimexpr\linewidth-0em]{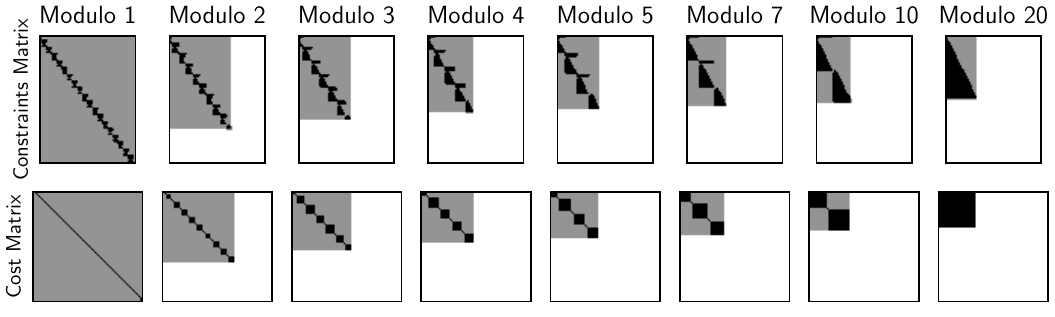}\null\vskip-.5ex
\caption{Sparsities of the Dual-Loop MPC with a Horizon of $N=20$ Steps
\\
The optimal solution to a quadratic program remains unchanged if we condense some optimization variables.
However, the corresponding constraint and cost matrices change in size (gray) and non-zero entries (black).
We use any number of blocks that can be generated for a given horizon.
The patterns vary from $N$ blocks for the sparse formulation to one block for the dense formulation.
}
\label{fig:sparsity}
\end{figure}

The dual-loop MPC~(\ref{eq:dual_loop_mpc}) is given in sparse form (multiple shooting).
Because different solvers perform best under different sparsity patterns,
we condensed the problem fully (single shooting) or partially for each horizon (Figure~\ref{fig:sparsity}).
For each horizon, we ran every solver with all relevant sparsity formulations.

\begin{figure}[b!]\centering
\!\includegraphics[width=\dimexpr\linewidth-0em]{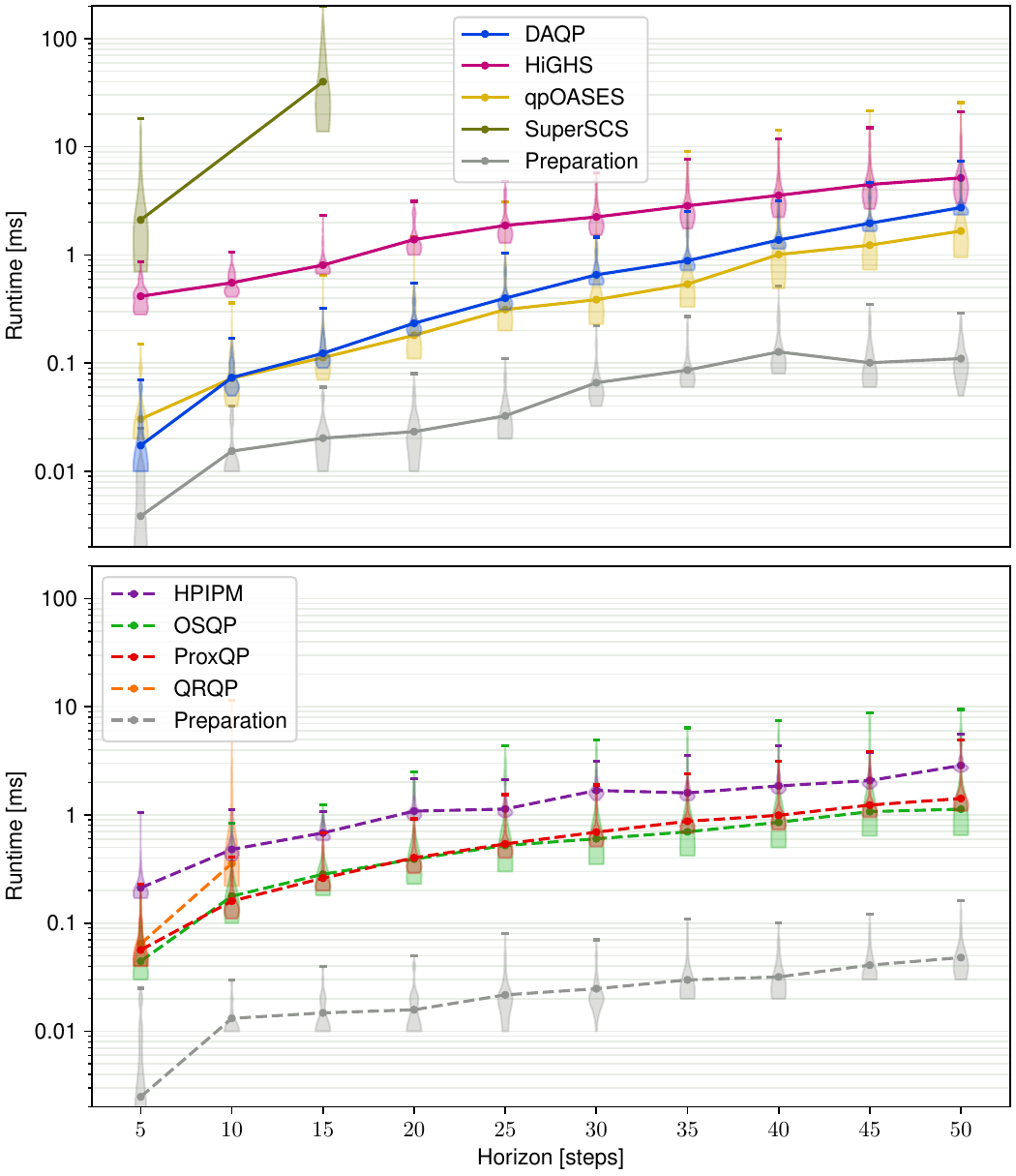}\null\vskip-1ex
\caption{Runtimes of Dual-Loop MPC with CasADi on Intel i7-11850H\\
For each solver and horizon, we selected the sparsity formulation that yields the lowest runtime.
We divided the solvers into two plots: those using condensed matrices (solid lines) and those using sparser matrices (dashed lines).
In each case, preparation accounts for only a fraction of the total runtime (note the logarithmic scale).
\\
DAQP and qpOASES are the fastest for smaller horizons.
Yet, the latter exhibits high variation in runtime;
computing the initial solution is particularly expensive.
ProxQP and OSQP are the fastest for larger horizons, though the runtimes of the former vary less.
All of them have lower mean runtimes than HPIPM and HiGHS.
However, the former exhibits low variation in runtime, and for larger horizons its maximum runtime is lower than that of qpOASES and OSQP.
No violin plots are shown for SuperSCS, QRQP, or ipqp at horizons for which they failed to solve all quadratic programs.}
\label{fig:casadi}
\end{figure}

We implemented the MPCs using CasADi~\cite{Andersson2018}. It can generate C code for the corresponding quadratic programs and interfaces the following open-source solvers:
DAQP~\cite{arnstrom2022dual},
HiGHS~\cite{Huangfu2018},
HPIPM~\cite{frison2020hpipmhighperformancequadraticprogramming},
ipqp,
OSQP~\cite{osqp},
ProxQP~\cite{bambade:hal-04133055},
qpOASES~\cite{Ferreau2014},
QRQP,
and
SuperSCS~\cite{superscs}.
Figure~\ref{fig:casadi} shows the runtimes of the solvers at different horizons.

\begin{figure}[thbp]\centering
\includegraphics[width=\dimexpr\linewidth]{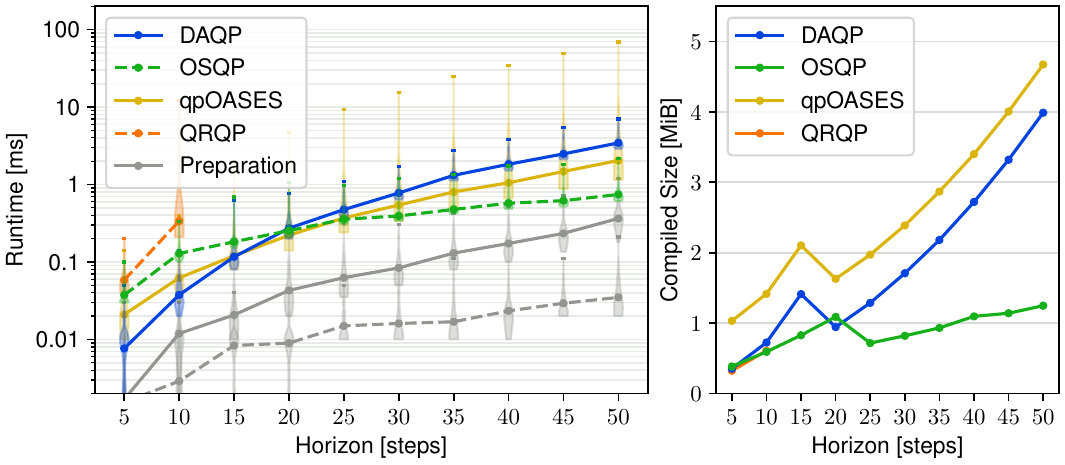}\vskip-1ex
\caption{Workload of Dual-Loop MPC in C++ on Intel i7-11850H\\
DAQP is the fastest solver for smaller horizons, where it can benefit from a dense formulation.
However, sparser formulations scale better, and OSQP is faster from 20 steps onwards.
The runtime of qpOASES falls between the two.
Its runtime is slightly higher only at 20 steps, albeit with high variance.
QRQP succeeded only for small horizons and was slower than the other solvers.
\\
The compiled size for DAQP and qpOASES grows much faster with the horizon than for OSQP.
This is particularly relevant for deployment on embedded devices with more restricted memory.}
\label{fig:mock}
\end{figure}

We identified the best-performing solvers and statically compiled C++ code to solve the generated quadratic programs with DAQP, OSQP, qpOASES, and QRQP. Figure~\ref{fig:mock} shows runtime behavior similar to that observed with CasADi for the selected solvers, except that OSQP is notably faster. In contrast, qpOASES is slightly slower due to the use of its integrated linear algebra backend, which is platform agnostic.

We analyzed the previous MPC in~\cite{Turnwald2025} for attitude pointing. Notably, QRQP performed well in that case. Here, however, it proves to be quite fragile because of the more demanding attitude tracking problem. It cannot complete a full run on the embedded device even for small horizons.

Preparing the quadratic programs to be solved is slower in the dense formulation than in the sparse formulations because it requires more matrix multiplications.
However, both preparations are minor in view of the solvers' runtimes.


\subsection{Delivery to an Embedded Platform}\label{sec:embedded}

We ported the MPCs with the selected solvers to the microcontroller Teensy 4.1 (ARM Cortex M-7) using CMake.
Initially, we tried the Arduino IDE and PlatformIO, which support the embedded device inherently.
The standalone code generated by CasADi (including QRQP) and OSQP~\cite{osqp-codegen} worked there immediately; yet, libraries for other solvers were not available.
Instead, we used their build files directly in CMake, included the library for the embedded platform and set up the parameters for the compilation.

On the ARM Cortex M-7, the runtimes are approximately 40 times higher than on the Intel i7-11850H in accordance with their clock speed.
Notably, for each horizon, the runtimes for the preparations vary much less.
This is because the preparations only involve a fixed number of matrix operations (without iterations) and the code is executed standalone on the embedded device.

\begin{figure}[thbp]\centering
\includegraphics[width=\dimexpr\linewidth]{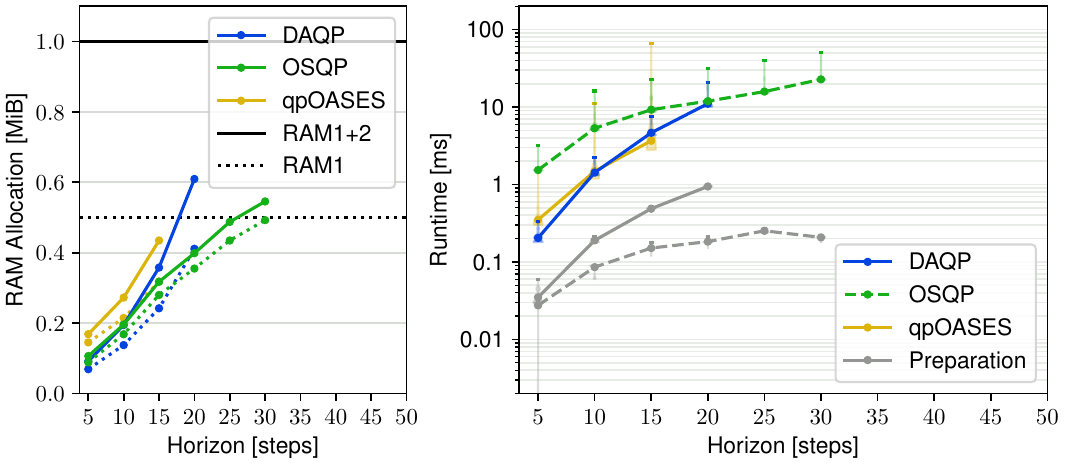}\vskip-1ex
\caption{Workload of Dual-Loop MPC in C++ on ARM Cortex M-7\\
For all selected solvers, the fast memory (RAM1) is the limiting resource for the horizon.
The memory consumption is about linear for OSQP, which benefits from a sparser formulation. In contrast, it increases more with the horizon for DAQP and qpOASES as they work with dense matrices.
\\
When it has enough memory, DAQP is best suited (up to 20 steps).
OSQP works up to 30 steps, where we would expect it to be the fastest solver, too.
However, the other solvers cannot run there:
QRQP could not solve all arising quadratic programs even for small horizons, and qpOASES runs out of memory after 15 steps.
At 15 steps, qpOASES has the lowest mean runtime but with high variation.}
\label{fig:teensy}
\end{figure}

The dual-loop MPC runs on the Teensy 4.1 for horizons up to 30 steps.
Figure~\ref{fig:teensy} shows that DAQP is best suited for horizons up to 20 steps, while only OSQP works for larger horizons.

The limiting factor is the available RAM.
For the generated CasADi code, we shift work arrays to the slower memory (DMAMEM).
It had a negligible impact on the runtimes, but the code could run for larger horizons.
There is room for improvement by shifting further allocations.

\section{\textsc{Conclusion}}

We developed two linear MPCs based on the multiplicative attitude deviation in reverse order, which makes the constraints on the spacecraft's angular velocity inherently linear.
Both exhibit accurate attitude tracking in the high-fidelity astrodynamics simulation Basilisk while obeying angular-velocity and torque constraints.
They run on a commercial off-the-shelf processor (ARM Cortex-M7), which is representative of a New Space platform, supporting onboard deployment for attitude tracking in resource-constrained spacecraft.

The dual-loop MPC outperforms the single-loop MPC in terms of runtime and memory consumption:
it is faster for smaller horizons while requiring less memory for larger horizons at similar runtimes.
This is because the dual-loop MPC is based on a linear time-invariant system, whereas the linear system of the single-loop MPC is parameter-varying.

Future work includes running the MPC on the embedded device in closed loop with the simulation.
To improve disturbance resilience, it could be extended to a robust MPC formulation, as in~\cite{Meinert2024}.
The controllers could consider relative deviations of spacecraft in a formation
for decentralized coordinated attitude control like in~\cite{VanDyke}.
The controllers could also be extended to the affine group $\mathrm{SE}(3)$ including translational motion as in~\cite{Teng2023}, e.g., for close proximity rendezvous with $6$ degrees of freedom.

\bibliographystyle{IEEEtran}
\bibliography{IEEEabrv,refs}

\end{document}